\documentclass[10pt, a4paper]{article}

\usepackage[review]{lrec-coling2024} 
\usepackage{times}
\usepackage{latexsym}
\usepackage{arabtex}
\usepackage{utf8}
\usepackage{graphicx}
\setcode{utf8}
\usepackage[T1]{fontenc}
\usepackage[utf8]{inputenc}
\usepackage{microtype}
\usepackage{inconsolata}
\usepackage{enumitem}
\usepackage{ragged2e}

\title{Creating Arabic LLM Prompts at Scale\\ \vspace*{.5\baselineskip} }

\name{Abdelrahman El-Sheikh, Ahmed Elmogtaba, Kareem Darwish, \\\large\textbf{Muhammad Elmallah, Ashraf Elneima, Hassan Sawaf}} 

\address{ aiXplain Inc. \\
         San Jose, CA, USA \\
         \{abdelrahman.el-sheikh,ahmed.abdelaziz,kareem.darwish,\\ muhammad.elmallah,ashraf.hatim,Hassan\}@aiXplain.com\\}

\abstract{
The debut of chatGPT and BARD has popularized instruction following text generation using LLMs, where a user can interrogate an LLM using natural language requests and obtain natural language answers that matches their requests. Training LLMs to respond in this manner requires a large number of worked out examples of user requests (aka prompts) with corresponding gold responses. In this paper, we introduce two methods for creating such prompts for Arabic cheaply and quickly.  The first methods entails automatically translating existing prompt datasets from English, such as PromptSource and Super-NaturalInstructions, and then using machine translation quality estimation to retain high quality translations only.  The second method involves creating natural language prompts on top of existing Arabic NLP datasets. Using these two methods we were able to create more than 67.4 million Arabic prompts that cover a variety of tasks including summarization, headline generation, grammar checking, open/closed question answering, creative writing, etc.  We show that fine tuning an open 7 billion parameter large language model, namely base Qwen2 7B, enables it to outperform a state-of-the-art 70 billion parameter instruction tuned model, namely Llama3 70B, in handling Arabic prompts. 
\\ \newline  }

\begin{document}
\maketitleabstract
\section{Introduction}
The debut of instruction-following Large Language Models (LLMs), such as chatGPT and BARD, showed that LLMs are capable of performing a variety of natural language tasks in response to user prompts. For example, a user may ask to summarize or extract specific information from a piece of text and an LLM would respond with the requested information.  

Tuning LLMs to behave in such manner requires using a large number of worked out examples of user requests (aka prompts) with corresponding correct gold responses. Many efforts have focused on creating such prompts in English resulting in datasets such PromptSource \cite{bach2022promptsource}, Super-NaturalInstuctions \cite{supernaturalinstructions,naturalinstructions}, FLAN \cite{wei2021finetuned}, Dolly v2 \cite{DatabricksBlog2023DollyV2}, and others.  Such efforts have resulted in the creation of tens of millions of prompts that cover a plethora of different tasks.

In this paper, we focus on creating high quality prompts at scale for Arabic, where work on prompt generation has been very limited.  We focus on two different methods for prompt creation.

The first method follows in the footsteps of PromptSource, where we created prompts for 76 existing Arabic Natural Language Processing (NLP) datasets using the PromptSource sourcing tool. The datasets include a variety of different tasks such as question answering, summarization, dialect identification, hate speech detection, diacritic recovery, grammar checking, and many more.  Using this method, we generated more than 67.4 million prompts.

The second method involves translating existing English prompt datasets with automated translation quality estimation and manual verification. Specifically, we translate both PromptSource and Super-NaturalInstructions, and we use state-of-the-art referenceless machine translation quality estimation \cite{rei2023scaling} to retain only prompts where all the sentences that make up the prompt have a minimum quality score. To ensure high quality prompts, we set a high minimum quality score, resulting in retaining roughly 20\% of the original prompts. Further, we perform manual verification and correction of translated prompts and manually inspect samples from each dataset to identify systematic translation errors.  Using this method, we created roughly 20 million prompts.  Both these methods can be applied to other languages.

We split the newly created prompts into train, validation, and test splits for all our datasets.  We use the test splits to score multiple existing LLMs. Further, we fine tune a strong LLM, namely Qwen2 7B, using roughly 10\% and 1\% of our newly created prompt datasets.  We show that the fine tuned model outperforms much larger models, such as Llama3-Instruct 70B.  Our testing mainly focuses on the ability of LLM to follow instructions in performing a variety of tasks as opposed to measuring their knowledge, which would typically be measured using standard test sets such as MMLU and Hellaswag.

The contribution of this paper is as follows:
\begin{itemize}[leftmargin=*]
    \item We create more than 87 million prompts for Arabic based on hundreds of datasets that cover 67 different tasks.
    \item We show that we can create high quality prompts using translation and translation quality estimation paired with manual verification.
    \item We show that fine tuning an LLM using the new data significantly improves LLM performance on a variety of tasks.
\end{itemize}


\section{Prompt Creation}

\subsection{Creating Prompts Using Existing Datasets}

\begin{figure}
    \centering
    \includegraphics[width=\linewidth]{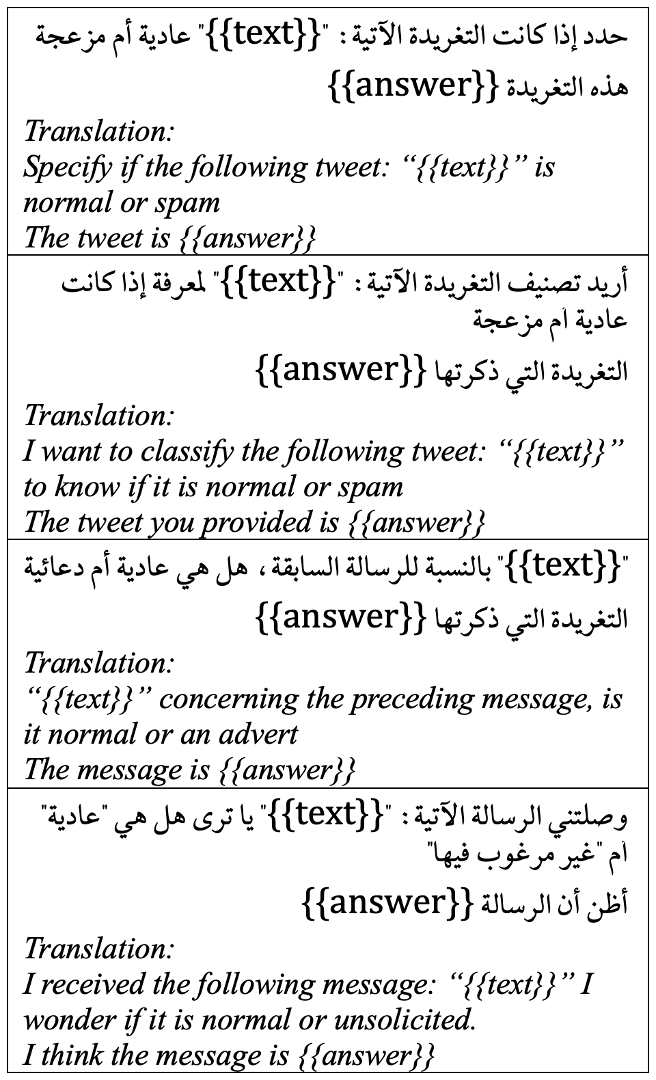}
    \caption{Example prompts for Abdelkareem/arabic\_tweets\_classification}
    \label{fig:prompt_examples}
\end{figure}

We created natural language prompts for 78 publicly available Arabic datasets using the PromptSource sourcing tool. The PromptSource UI facilitates a user-friendly interface for the creation of prompt templates, providing the ability to specify the insertion points of inputs within the prompts and to provide potential answers.  For each task we developed 2-8 different prompt templates.  The time required for creating the prompt templates was typically 30-90 minutes.  Figure \ref{fig:prompt_examples} shows 4 different prompt templates that we developed for one of the datasets\footnote{\url{https://huggingface.co/datasets/Abdelkareem/arabic\_tweets\_classification}}.  We were keen to diversify the order and language of the prompts.  From 78 datasets, we generated 67,488,303 prompts.

\subsection{Creating Prompts Using Translation}

We developed a comprehensive data processing pipeline designed to perform the translation, evaluation, and filtering of English datasets into Arabic. The pipeline consists of a series of well-defined steps, summarized as follows:
\begin{itemize}
    \item \textbf{Sentence Extraction:} Initially, the pipeline splits the English prompts into sentences.  We used the Python sentence-splitter library\footnote{\url{https://github.com/mediacloud/sentence-splitter}}, which is a re-implementation of the Moses sentence splitter\footnote{\url{https://www.statmt.org/europarl/}}. These sentences serve as the source text for translation.
    \item \textbf{Machine Translation:} Subsequently, we translated English sentences to Arabic using the opus-mt-ar-en Helsinki Machine Translation (MT) model \cite{tiedemann-thottingal-2020-opus}, which is a neural model based on Marian NMT\footnote{\url{https://marian-nmt.github.io}}.  We opted to use this model as opposed to commercial translation APIs to lower the overall cost.
    \item \textbf{Translation Evaluation:} In the pursuit of ensuring translation quality, the pipeline incorporates the COMET-QE reference-free MT evaluation model \cite{rei2023scaling}, which has been shown to correlate well with human judgments. COMET-QE calculates a quality score on a scale from 0 to 1, with 0 indicating the lowest translation quality and 1 signifying the highest quality.
    \item \textbf{Threshold-Based Filtering:} To enhance the reliability of the translated content, a specified threshold, namely 0.7, is applied. A translated prompt  where one or more of its sentences do not meet this quality criterion are filtered out, ensuring that only high-quality translated prompts are retained for further utilization.
    \item \textbf{Manual Translation Verification:} We manually correct the translation of the instructions for datasets where the instructions were separable from the input.
    \item \textbf{Quality Assurance:} We manually check a random sample from each dataset to determine the efficacy of translation. Performing this manual check helped us identify (and subsequently drop) datasets that are not amenable for translation such as those that require the identification of English specific grammatical phenomena.  It also helped us in identifying systematic translation errors that can be easily handled.  For example, some datasets involved the manipulation of a list of numbers, and thus aside from the instructions, the input and output did not require translation.  Another example involves the incorrect translation of word ``passage'', which was translated to mean a ``walk way'' instead of a ``piece of text''.
\end{itemize}

Our target was to translate both PromptSource and Super-NaturalInstructions, which are composed of 57 and 1,429 datasets respectively. Prior to translation, we excluded datasets that were English specific such as grammar checking and translation tasks such translation between Urdu and Marathi.  In doing so, we excluded 14 and 490 tasks from PromptSource and Super-NaturalInstructions respectively.  After running our pipeline on both datasets, we excluded tasks that had less than 10 translated prompts after filtering.  While none of the tasks for PromptSource were excluded, an additional 619 tasks were excluded for Super-NaturalInstructions. 

\begin{tabular}{l|r|r}
     Source & Datasets & Prompts \\ \hline
     PromptSource & 43 & 19,555,120 \\
     Super-NaturalInstruct & 320 & 345,168 \\
\end{tabular}

\section{Fine Tuning Using New Data}

\subsection{Data Splits}

From all the prompts that we created, we extracted a test set composed of a random sample from each dataset with a maximum of 50 samples from each dataset.  The final size of the test set was 13,462 prompts with ground truth responses that was extracted from 643 datasets that cover 67 different tasks.  For fine tuning, we constructed two training datasets composed of 800 thousand prompts and 8 million prompts, which were randomly extracted from all the datasets (and were not in the test set).  We opted to create two training datasets to measure the impact of using more (or less) examples for fine tuning.  From the training dataset, we randomly picked 1\% of the prompts for validation.

\subsection{Baselines}

As baselines, we prompted a variety of LLMs using our test set. 
 The models were:
\begin{itemize}
    \item \textbf{AceGPT-Instruct 7B} \cite{huang2023acegpt}: This is one of the first publicly available LLMs that focus on Arabic. The model is based on Llama2, where the model was further pre-trained using Arabic text and instruction tuned using Arabic instruction datasets.
    \item \textbf{Jais-chat 13B} \cite{sengupta2023jais}: This is the first open Arabic/English LLM that was trained from scratch based on the GPT-3 decoder architecture.
    \item \textbf{Llama3-Instruct 8B \& 70B} \cite{dubey2024llama}: These are large foundational models from Meta. Though they are primarily pre-trained on English text, the models have demonstrated reasonable instruction following abilities for other languages.
    \item \textbf{Qwen2-Instruct 7B} \cite{yang2024qwen2}: These are large foundational models that are multilingual by design and have achieved competitive results on standard benchmarks.  We later fine tune the base Qwen2 7B model using our instruction data.
\end{itemize}

We used ROUGE-L as our quality metric. Table \ref{tab:results} shows the baseline results across all the aforementioned models for a variety of tasks.  The results show that Llama3 70B achieved the best results followed by Llama3 8B, and AceGPT 7B yielded the worst results.  

\subsection{Fine Tuning Setup}

We fine tuned the base Qwen2 7 billion parameter model using Low-Rank Adaptation (LoRA) \cite{hu2021lora}. This is a parameter-efficient method that freezes the parameters in the existing layers of a model and introduces trainable rank decomposition matrices for each layer in the model, reducing the parameters that require tuning by orders of magnitude, leading to significantly smaller memory requirements while achieving results that are at par with full fine tuning. We used the peft library from HuggingFace\footnote{\url{https://huggingface.co/docs/peft/en/index}}.

We fine tuned Qwen2 7B using two different sets of instructions that we randomly sampled from our newly created instruction datasets.  The size of the first sample was 800K, and the second was 8M. We chose two different sample sizes to quantify the effect of using increased training data.  
We used the following Arabic system prompt and prompt structure for all our training instruction:

\<فيما يلي تعليمات باللغة العربية تصف المهمة.> \<  اكتب ردًا باللغة العربية يكمل الطلب بشكل مناسب.>

(Translation: ``The following are Arabic instructions that describe a task.  Provide an answer in Arabic that satisfies the request

\#\#\# Instruction:

\{text of the instruction\}

\#\#\# Response:

\{text of the response\}

We fine tuned the model for one epoch and the key tuning hyper parameters were as follows:
\begin{itemize}
    \item \textbf{Warmup Ratio:} We used a warm-up ratio of 0.1, meaning that the learning rate gradually increased from a low value to its initial value over the first 10\% of the training steps. This helps stabilize the training process.
    \item \textbf{Learning Rate:} The learning rate was set to 2e-4 and adjusted using a linear scheduler, which decreases the learning rate steadily during training. This helps the model converge smoothly.
    \item \textbf{Optimizer:} We used the Adam optimizer, which is effective in handling sparse gradients and changes in the training data. This choice, combined with our learning rate settings, ensures stable and effective updates to the model's parameters.
    \item \textbf{Weight Decay:} We also used a weight decay of 0.01 to prevent overfitting and to improve the generalization of the model.
\end{itemize}

\subsection{Fine Tuning Results}

As shown in Table \ref{tab:results}, fine tuning Qwen2 7B model using our carefully curated dataset of 800k samples led to statistically significant improvement over the Qwen2-Instruct 7B with respective average ROUGE-L scores of 0.184 and 0.143.  We used a paired 2-tailed t-test over per dataset scores with a p-value less than 0.01 to determine statistical significance. In turn, fine tuning using 8M samples statistically significantly outperformed all baseline models, except for Llama 3-70B (ROUGE-L = 0.221; p-value > 0.79), and Qwen2 7B fine tuned on 800k examples with an average ROUGE-L score of 0.224. It performed better than the Arabic-focused models AceGPT 7B and Jais 13B, which scored average ROUGE-L scores of 0.102 and 0.138, respectively. This demonstrates the effectiveness of fine tuning using our dataset with significant improvement across a variety of tasks, where the fine tuned Qwen2 7B 800k and 8M models respectively performed 29\% and 57\% higher than the Qwen2 7B-Instruct model.
We also observe that fine tuning using more examples (8M versus 800k) leads to 22\% improvement in results. This suggests that  fine tuning using more examples positively impacts the final model's performance.



\begin{table*}[]
    \centering
    \scriptsize
    \begin{tabular}{l|r|p{0.95cm}|p{0.95cm}|p{0.95cm}|p{0.95cm}|p{0.95cm}||p{0.95cm}|p{0.95cm}}
Task & No. of Datasets & AceGPT 7B & Jais-chat 13B & Llama 3 8B & Llama 3 70B & Qwen2 7B-Instruct & Qwen2 7B-FT-800k & Qwen2 7B-FT-8M \\ \hline
Answer Verification & 1 & 0.014 & 0.073 & 0.082 & 0.075 & 0.680 & 0.600 & \textbf{0.667}\\ 
Answerability Classification & 2 & 0.041 & 0.100 & 0.108 & 0.105 & 0.362 & 0.467 & \textbf{0.613}\\ 
Cause Effect Classification & 3 & 0.091 & 0.183 & 0.140 & \textbf{0.200} & 0.070 & 0.032 & 0.135\\ 
Classification & 18 & 0.098 & 0.151 & 0.147 & \textbf{0.216} & 0.061 & 0.168 & 0.187\\ 
Coherence Classification & 2 & 0.046 & 0.081 & 0.048 & 0.051 & 0.800 & 0.402 & \textbf{0.560}\\ 
Command Interpretation & 1 & 0.036 & 0.103 & 0.114 & \textbf{0.116} & 0.052 & 0.026 & 0.035\\ 
Commonsense Validation & 2 & 0.046 & 0.131 & 0.087 & 0.059 & 0.077 & 0.088 & \textbf{0.154}\\ 
Coreference Resolution & 3 & 0.109 & 0.092 & \textbf{0.171} & 0.163 & 0.104 & 0.024 & 0.036\\ 
Data to Text & 2 & 0.073 & 0.099 & 0.099 & \textbf{0.172} & 0.127 & 0.067 & 0.115\\ 
Definition Generation & 1 & 0.037 & \textbf{0.138} & 0.128 & 0.120 & 0.090 & 0.035 & 0.010\\ 
Diacritization & 15 & 0.535 & 0.315 & 0.563 & \textbf{0.802} & 0.277 & 0.265 & 0.399\\ 
Dialect Classification & 6 & 0.074 & 0.152 & \textbf{0.277} & 0.229 & 0.080 & 0.105 & 0.097\\ 
Dialogue Generation & 11 & 0.073 & 0.118 & 0.116 & 0.117 & 0.071 & 0.132 & \textbf{0.142}\\ 
Disease Mention Identification & 1 & 0.030 & 0.097 & 0.247 & 0.243 & 0.028 & \textbf{0.500} & 0.109\\ 
Duplicate Question Identification & 2 & 0.015 & 0.108 & 0.041 & 0.078 & 0.009 & \textbf{0.750} & 0.113\\ 
Emotion Detection & 12 & 0.038 & 0.044 & 0.071 & 0.118 & 0.022 & \textbf{0.275} & 0.183\\ 
Ethics Classification & 2 & 0.081 & 0.124 & 0.095 & 0.085 & 0.011 & 0.381 & \textbf{0.440}\\ 
Explanation & 6 & 0.145 & 0.161 & 0.198 & \textbf{0.209} & 0.104 & 0.086 & 0.173\\ 
Fact Verification & 5 & 0.055 & 0.121 & 0.155 & 0.201 & 0.123 & 0.380 & \textbf{0.441}\\ 
Fill in The Blank & 3 & 0.047 & 0.123 & 0.162 & \textbf{0.171} & 0.025 & 0.144 & 0.081\\ 
Grammar Correction & 14 & 0.293 & 0.258 & 0.347 & \textbf{0.514} & 0.416 & 0.359 & 0.377\\ 
Inference Detection & 1 & 0.024 & 0.194 & 0.308 & 0.381 & 0.022 & \textbf{0.900} & \textbf{0.900}\\ 
Information Extraction & 21 & 0.083 & 0.092 & 0.207 & \textbf{0.211} & 0.140 & 0.132 & 0.179\\ 
Information Retrieval & 1 & 0.055 & 0.083 & 0.071 & 0.096 & 0.113 & 0.125 & \textbf{0.173}\\ 
Instruction Generation & 1 & 0.042 & \textbf{0.115} & 0.086 & 0.097 & 0.019 & 0.067 & 0.049\\ 
Intent Identification & 3 & 0.053 & \textbf{0.109} & 0.084 & 0.071 & 0.084 & 0.042 & 0.101\\ 
Keyword Extraction & 10 & 0.066 & 0.092 & 0.138 & \textbf{0.148} & 0.027 & 0.095 & 0.118\\ 
Logical Reasoning & 3 & 0.071 & 0.157 & 0.546 & 0.235 & 0.011 & \textbf{0.824} & 0.747\\ 
Misc. & 8 & 0.066 & 0.116 & 0.133 & \textbf{0.138} & 0.052 & 0.038 & 0.075\\ 
Named Entity Recognition & 14 & 0.105 & 0.133 & \textbf{0.261} & 0.272 & 0.193 & 0.054 & 0.097\\ 
Natural Language Inference & 1 & 0.090 & 0.203 & 0.689 & 0.509 & 0.647 & \textbf{0.720} & 0.480\\ 
News Article Generation & 2 & 0.123 & 0.107 & 0.158 & \textbf{0.175} & 0.142 & 0.105 & 0.074\\ 
Next Action Prediction & 1 & 0.067 & 0.103 & \textbf{0.220} & 0.204 & 0.042 & 0.062 & 0.096\\ 
Offensive Language Detection & 15 & 0.061 & 0.113 & 0.163 & 0.227 & 0.280 & \textbf{0.515} & 0.502\\ 
Paraphrasing & 9 & 0.175 & 0.160 & 0.254 & \textbf{0.264} & 0.216 & 0.123 & 0.162\\ 
Poem Generation & 1 & 0.092 & 0.092 & 0.122 & \textbf{0.132} & 0.060 & 0.025 & 0.076\\ 
Problem Identification & 1 & 0.021 & \textbf{0.096} & 0.035 & 0.022 & 0.020 & 0.001 & 0.078\\ 
Program Execution & 18 & 0.065 & 0.113 & 0.058 & 0.057 & 0.196 & 0.374 & \textbf{0.833}\\ 
Query Classification & 1 & 0.014 & 0.141 & 0.073 & 0.056 & 0.003 & \textbf{0.500} & 0.300\\ 
Question Answering & 115 & 0.083 & 0.148 & 0.194 & \textbf{0.235} & 0.130 & 0.149 & 0.150\\ 
Question Categorization & 1 & 0.075 & 0.081 & 0.253 & \textbf{0.334} & 0.000 & 0.222 & 0.073\\ 
Question Decomposition & 2 & 0.128 & 0.123 & 0.128 & 0.143 & \textbf{0.163} & 0.027 & 0.042\\ 
Question Generation & 43 & 0.085 & 0.117 & 0.145 & \textbf{0.149} & 0.153 & 0.068 & 0.106\\ 
Question Understanding & 3 & 0.095 & 0.157 & 0.111 & 0.149 & \textbf{0.365} & 0.293 & 0.213\\ 
Reading Comprehension & 10 & 0.081 & 0.198 & 0.321 & 0.359 & 0.073 & 0.560 & \textbf{0.645}\\ 
Review Rating Prediction & 3 & 0.122 & 0.107 & 0.234 & 0.295 & 0.046 & \textbf{0.154} & 0.121\\ 
Riddle Solving & 1 & 0.036 & 0.105 & 0.121 & \textbf{0.165} & 0.034 & 0.000 & 0.003\\ 
Sarcasm Detection & 7 & 0.094 & 0.176 & 0.194 & 0.264 & 0.081 & 0.411 & \textbf{0.499}\\ 
Semantic Similarity & 21 & 0.087 & 0.205 & 0.185 & 0.189 & 0.209 & 0.382 & \textbf{0.497}\\ 
Sentence Composition & 11 & 0.156 & 0.155 & 0.190 & \textbf{0.195} & 0.172 & 0.050 & 0.085\\ 
Sentence Ordering & 1 & 0.051 & 0.082 & 0.161 & 0.164 & \textbf{0.348} & 0.098 & 0.216\\ 
Sentiment Analysis & 41 & 0.110 & 0.159 & \textbf{0.273} & 0.240 & 0.129 & 0.235 & 0.229\\ 
Spam Detection & 4 & 0.090 & 0.069 & 0.084 & 0.138 & 0.025 & 0.232 & \textbf{0.406}\\ 
Stereotype Detection & 5 & 0.111 & 0.163 & 0.388 & 0.415 & 0.227 & 0.269 & \textbf{0.480}\\ 
Story Composition & 7 & 0.104 & 0.112 & \textbf{0.167} & 0.179 & 0.075 & 0.041 & 0.043\\ 
Style Transfer & 1 & 0.097 & 0.096 & 0.157 & \textbf{0.158} & 0.040 & 0.001 & 0.044\\ 
Summarization & 36 & 0.111 & 0.141 & 0.182 & \textbf{0.198} & 0.154 & 0.078 & 0.122\\ 
Text Categorization & 7 & 0.038 & 0.094 & 0.139 & 0.161 & 0.244 & 0.306 & \textbf{0.491}\\ 
Text Classification & 4 & 0.058 & 0.128 & 0.204 & 0.324 & 0.100 & 0.324 & \textbf{0.575}\\ 
Text Completion & 9 & 0.084 & 0.116 & 0.166 & 0.111 & 0.028 & 0.163 & \textbf{0.270}\\ 
Text Generation & 8 & 0.059 & 0.062 & \textbf{0.078} & 0.076 & 0.070 & 0.050 & 0.044\\ 
Text Simplification & 2 & 0.222 & 0.193 & 0.323 & \textbf{0.353} & 0.287 & 0.091 & 0.149\\ 
Title Generation & 28 & 0.082 & 0.119 & 0.162 & \textbf{0.184} & 0.118 & 0.079 & 0.075\\ 
Topic Identification & 5 & 0.076 & 0.098 & 0.171 & \textbf{0.293} & 0.043 & 0.101 & 0.165\\ 
Translation & 22 & 0.108 & 0.095 & 0.175 & 0.245 & 0.185 & 0.138 & \textbf{0.203}\\ 
Transliteration & 10 & 0.102 & 0.130 & 0.148 & \textbf{0.228} & 0.086 & 0.079 & 0.122\\ 
Wrong Candidate Generation & 13 & 0.064 & 0.102 & 0.105 & \textbf{0.114} & 0.054 & 0.062 & 0.027\\ 
\hline
\multicolumn{2}{r}{AVERAGE}	&	0.102	&	0.138	&	0.192	&	0.221	&	0.143 & 0.184	&	\textbf{0.224}	\\
    \end{tabular}
    \caption{Results for baseline models and fine tuned Qwen2 models.}
    \label{tab:results}
\end{table*}

\section{Conclusion}
In this paper, we presented two robust methods for creating large prompt datasets for Arabic.  The first method involves creating prompts based on applying prompt templates.  The second method involves automatically translating existing English prompt datasets, applying automatic filtering of translations using reference-free quality estimation and manual verification.  In all, we created more than 87 million prompts.  We fine tuned a state-of-the-art 7 billion parameter foundational model, namely Qwen2 7B, using roughly 1\% and and 10\% of the prompts that we have created, and show that both fine tuned models statistically significantly outperform the publicaly available instruction tuned version of the model.  Further, our fine tuned model using 10\% of our prompt data slightly outperforms a much larger instruction tuned 70 billion parameter model, namely Llama3 70B.  We show that the performance of the fine tuned models positively correlates with fine tuning using larger prompt datasets.


\section{{References}}
\bibliography{references, custom, sample}
\bibliographystyle{lrec_natbib}
\end{document}